\pdfoutput=1
\documentclass[11pt]{article}

\usepackage{ACL2023}
\usepackage{times}
\usepackage{latexsym}
\usepackage{inconsolata}
\usepackage{amsmath}
\usepackage{amsfonts}
\usepackage{amssymb}
\usepackage{graphicx}
\usepackage{multirow}
\usepackage{booktabs}
\usepackage{diagbox}
\usepackage{bbding}
\usepackage{algpseudocode}
\usepackage{enumitem}
\usepackage{commath}
\usepackage{algorithm}
\usepackage[toc,page]{appendix}
\usepackage{tabularx}
\usepackage{listings}
\usepackage{colortbl}
\usepackage{xcolor}
\usepackage{comment}
\usepackage{hyperref}
\usepackage[T1]{fontenc}
\usepackage[utf8]{inputenc}

\usepackage{microtype}
\usepackage{inconsolata}

\usepackage[colorinlistoftodos]{todonotes}

\title{\textit{SafeSpeech}: A Comprehensive and Interactive Tool for \\ Analysing Sexist and Abusive Language in Conversations}

\definecolor{rowcolor}{gray}{0.9}

\author{Xingwei Tan$^{1,2}$, Chen Lyu$^1$, Hafiz M. Umer$^1$, Sahrish Khan$^1$, Mahathi Parvatham$^1$, \\
\textbf{Lois Arthurs}$^{4}$, \textbf{Simon Cullen}$^{4}$, \textbf{Shelley Wilson}$^{4}$, \textbf{Arshad Jhumka}$^{1,3}$, \textbf{Gabriele Pergola}$^1$ \\[0.3em]
  $^1$Department of Computer Science, University of Warwick, UK\\
  $^2$School of Computer Science, University of Sheffield, UK\\
  $^3$School of Computer Science, University of Leeds, UK\\
  $^4$Forensic Capability Network, UK\\[0.3em]
  \texttt{\{xingwei.tan, chen.lyu, sahrish.khan, gabriele.pergola.1\}@warwick.ac.uk}\\
  \texttt{\{lois.arthurs, simon.cullen, shelley.wilson\}@dorset.pnn.police.uk}\\
  \texttt{h.a.jhumka@leeds.ac.uk}
}

\begin{document}

\maketitle
\begin{abstract}
Detecting toxic language including sexism, harassment and abusive behaviour, remains a critical challenge, particularly in its subtle and context-dependent forms. Existing approaches largely focus on isolated message-level classification, overlooking toxicity that emerges across conversational contexts. To promote and enable future research in this direction, we introduce \textit{SafeSpeech}, a comprehensive platform for toxic content detection and analysis that bridges message-level and conversation-level insights. The platform integrates fine-tuned classifiers and large language models (LLMs) to enable multi-granularity detection, toxic-aware conversation summarization, and persona profiling. \textit{SafeSpeech} also incorporates explainability mechanisms, such as perplexity gain analysis, to highlight the linguistic elements driving predictions. Evaluations on benchmark datasets, including EDOS, OffensEval, and HatEval, demonstrate the reproduction of state-of-the-art performance across multiple tasks, including fine-grained sexism detection.\footnote{\href{https://www.safespeechs.org/}{SafeSpeech Demo Website}}
\end{abstract}

\section{Introduction}

The dissemination of toxic content has become a pervasive issue across digital communication platforms. Toxic language, including sexism, harassment, and abusive behaviour, is not only widespread but also often amplified by the ease of dissemination in online environments. While explicit forms of harmful language, such as direct insults or threats, have been the focus of much research \cite{zampieri-etal-2019-predicting, vidgen-etal-2021-introducing}, there is increasing awareness that toxicity frequently manifests in more subtle or context-dependent forms. 
Subtle toxic behaviours include coercive language, controlling speech, and microaggressions  \cite{wijanarko-etal-2024-monitoring}. These may seem benign in isolation but reveal their toxicity through recurring patterns in conversations, as observed in contexts such as violence against women and girls (VAWG).
Recent advances, such as the EDOS dataset for fine-grained sexism detection \cite{kirk-etal-2023-semeval} and large language models (LLMs) capable of analyzing multi-turn dialogues \cite{suhara-alikaniotis-2024-source}, highlight the potential for bridging this gap.
Yet, existing approaches to toxic content detection mainly operate at the message level \cite{basile-etal-2019-semeval,caselli-etal-2020-feel,Mathew2020HateXplainAB}, using classifiers trained to identify the harmful language in isolated text \cite{camacho-collados-etal-2022-tweetnlp, wijanarko-etal-2024-monitoring}.
And to date, there is no comprehensive platform that integrates message-level classification with conversation-level analysis to systematically evaluate and benchmark model performance across these dimensions.

Therefore, we introduce \textit{SafeSpeech}, a unified platform designed for toxic content detection and analysis. \textit{SafeSpeech} bridges the gap between message-level classification and conversation-level analysis, providing users with the tools to evaluate both explicit and context-dependent forms of toxicity. 
The platform supports a broad range of tasks, from granular single-message classification to the contextual understanding of multi-turn conversations. It leverages recent advances in dataset availability \cite{kirk-etal-2023-semeval, basile-etal-2019-semeval, zampieri-etal-2019-semeval, caselli-etal-2020-feel} and large language models (LLMs) capable of processing extended sequences and capturing conversational context.

\textit{SafeSpeech} incorporates prompt-based customization, enabling users to apply tailored \textit{prompt templates} for classification tasks. 
To better understand the models' prediction, \textit{SafeSpeech} integrates a soft-explainability mechanism. This module, based on perplexity gain analysis \cite{suhara-alikaniotis-2024-source}, identifies and visually highlights the phrases that most contributed to a classification output. 

The platform extends its capabilities with a toxic-aware summarization module designed for long, multi-turn conversations. By combining semantic chunking with instruction-based summarization \cite{wang-etal-2023-instructive}, this module identifies recurring or overlapping topics, ensuring that semantically related content is grouped together, even when interrupted by unrelated dialogue, and produces concise summaries. These summaries are conditioned on toxic classification outputs, to explicitly highlight harmful content.
In addition to summarization, \textit{SafeSpeech} includes a persona analysis module that goes beyond detecting toxic content to characterize individuals’ behavioural tendencies. Existing methodologies offering automated persona analysis for conversations typically provide only general predictions \cite{han2023speaker, jcl132_personality_prediction, lyu24, wen2024affective}. In contrast, our platform represents the first approach to leverage the Big Five Personality Traits framework \cite{costa2008revised} and toxic-aware summaries to provide trait-based insights conditioned on toxic content classifications. 

\textit{SafeSpeech} has been thoroughly evaluated across diverse datasets and practical use cases. It provides state-of-the-art model performance within benchmark tasks for sexism, abusive, and hate speech detection, along with advanced tools for conversation summarization and persona analysis.
The contributions of this paper can be summarized as follows:

\begin{enumerate}
    \item \textit{SafeSpeech}, an integrated platform combining message-level and conversation-level analysis, provides state-of-the-art classifiers and datasets for fine-grained and context-dependent forms of toxicity.

    \item  An AI assistant for dynamic exploration of the classified data via prompt-based interactions, and a soft-interpretability mechanism based on perplexity gain, enabling users to better trace model predictions.

    \item  Modules for toxic-aware conversation summarization and for persona analysis, facilitating in-depth examination of speaker behaviour and dynamics within multi-turn dialogues.

    \item A comprehensive evaluation across benchmark datasets, demonstrating its state-of-the-art performance in identifying and analyzing nuanced toxic behaviours on public datasets.
\end{enumerate}

\section{Related Work}
Existing approaches to toxic content detection mainly operate at the message level \cite{basile-etal-2019-semeval, caselli-etal-2020-feel, Mathew2020HateXplainAB,pergola-etal-2021-disentangled, tan-etal-2023-event, tan-etal-2024-set, lyu-pergola-2024-scigispy, sun-etal-2024-leveraging}, which use classifiers to identify the harmful language in the isolated text without considering the complicated interactions in conversation settings.
These systems commonly framed toxic content detection as sentence classification, then followed by extensive feature engineering and supervised training of statistical machine learning models or deep neural networks \cite{DBLP:journals/semweb/ZhangL19, zhu-etal-2021-topic}.
Perspective API \cite{10.1145/3534678.3539147} is based on a BERT model pre-trained on ``subword'' sequences (UTF-8 bytes) and thus is capable of multilingual hate speech detection.
\citet{camacho-collados-etal-2022-tweetnlp} provide a comprehensive analysis platform for tweets with language identification modules for hate and offensive language detection, but can only produce binary labels on sentence-level.
\citet{wijanarko-etal-2024-monitoring} presents a system that can detect hate speech and identify its $5$ subtypes, even though they only provide IndoBERTweet \cite{koto-etal-2021-indobertweet}  as the backbone, which is trained only on Indonesia toxic speech data \cite{susanto2024indotoxic2024demographicallyenricheddatasethate}.

%%%%%%%%%%%%%%%%%%%%%%%%%%%%%%%%%%%%%
% System Description and Architecture
%%%%%%%%%%%%%%%%%%%%%%%%%%%%%%%%%%%%%
\section{\textit{SafeSpeech} Platform}

\subsection{Platform Architecture Overview}

\begin{figure}
    \centering
    \includegraphics[width=\columnwidth]{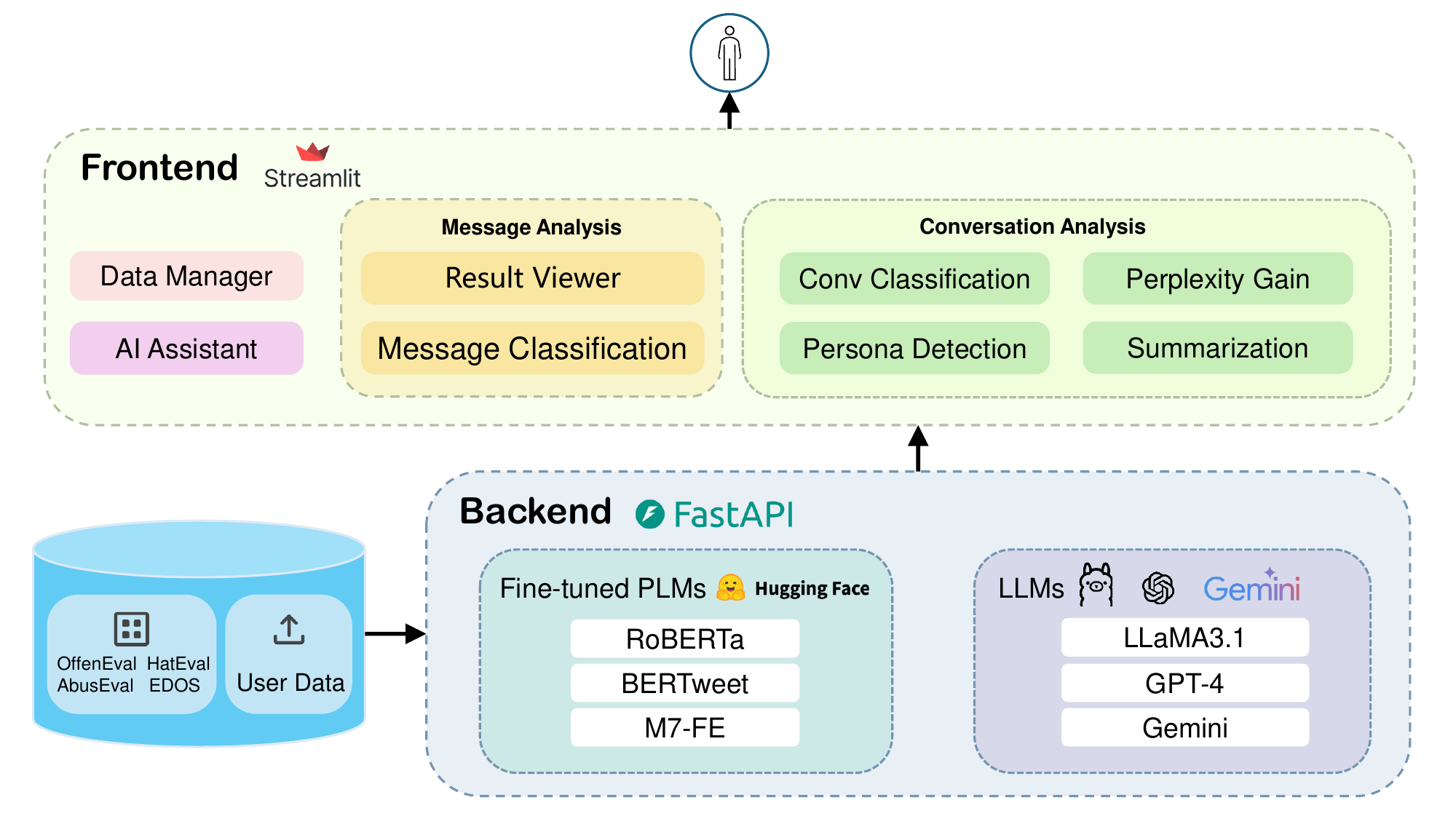}
    \caption{The structure of \textit{SafeSpeech} platform.}
    \label{fig:tool_structure}
\end{figure}

\textit{SafeSpeech} is designed as an interactive platform that combines data sources with analysis models, including neural classifiers and large language models (Figure \ref{fig:tool_structure}). 
Four core modules connect the data, analysis methods, and interpretation tools together:

\begin{figure*}[ht]
    \centering
    \includegraphics[width=0.85\linewidth]{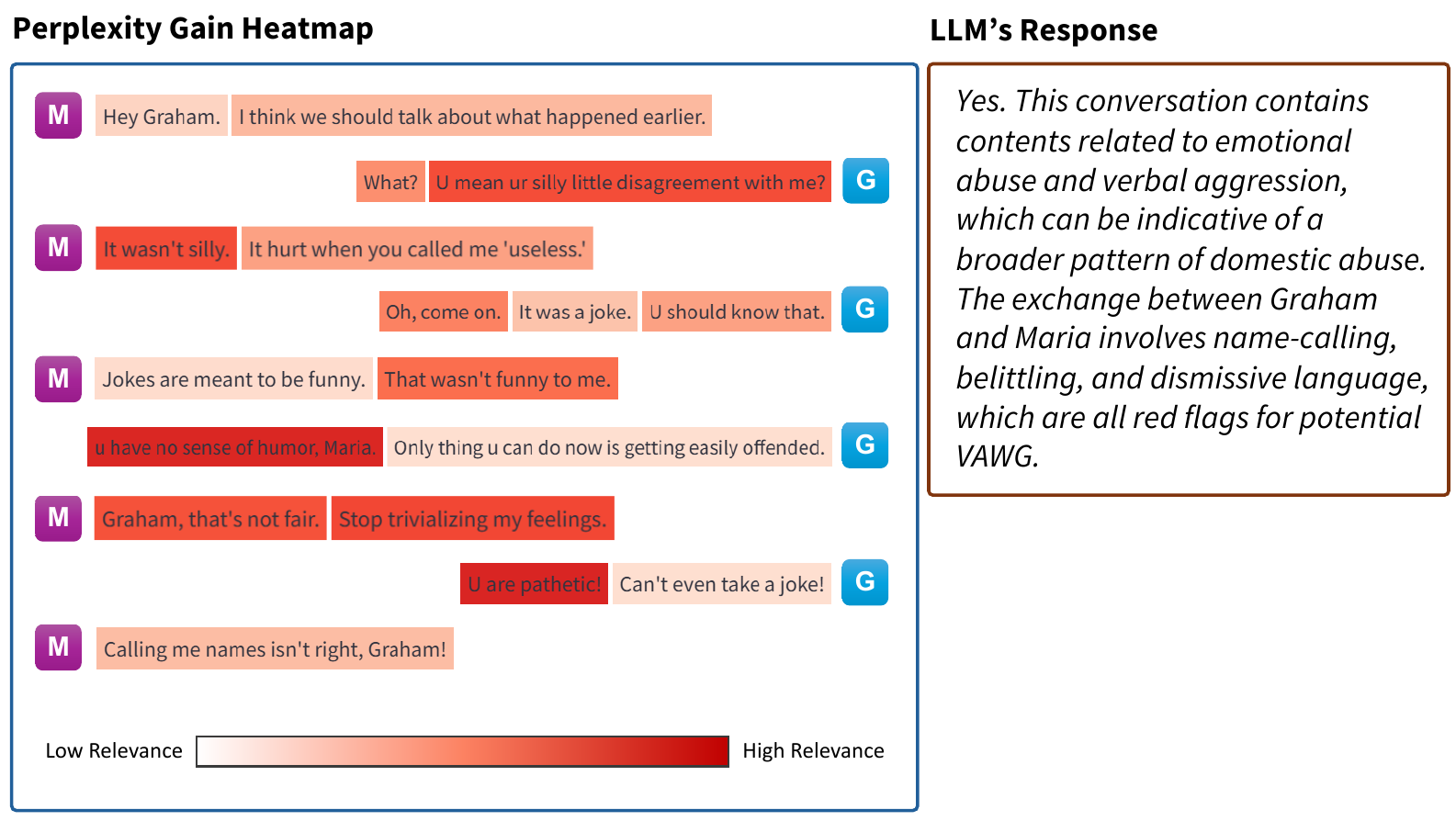}
    \caption{An example of perplexity gain analysis on a synthetically generated conversation. Based on the response, we computed perplexity gains and represented them as a heatmap.}
    \label{fig:ppl_example}
\end{figure*}

\begin{itemize}
    \item \textbf{Data Management} Host multiple toxic speech detection benchmark datasets. Manage user-uploaded data.
    \item \textbf{AI Assistant} Provide an LLM-based conversation interface, which is aided with carefully-designed prompts and classification labels, for interactive detection of toxic content. Users can also input custom prompts for detection and ask follow-up questions.
    \item \textbf{Message Analysis} Provide state-of-the-art fine-tuned classifiers on various toxic content detection subtypes, returning message-level predictions.
    \item \textbf{Conversation Analysis} Utilize LLMs for conversation-level toxic content detection. On top of the prediction results, we 1) provide perplexity gain to trace the source of the predictions, 2) label-aware conversation summarization, and 3) persona detection.
\end{itemize}

\subsection{User Interaction Design}

\textit{SafeSpeech} is an interactive, multi-functional platform for the comprehensive analyses of toxic speech in messages and conversations. The frontend is developed using \textit{Streamlit}\footnote{\href{https://streamlit.io/}{Streamlit}}, providing an intuitive, browser-based interface that requires no additional software installation. The main modules are available via a dedicated sidebar and allow the user to (i) upload and manage datasets, (ii) explore the results interactively via an AI assistant, and perform fine-grained analyses at both (iii) message and (iv) conversation levels (see Appendix \ref{appendix:screenshoots} for screenshots).

\subsubsection{Data Manager} Users can upload their data files or select from available benchmarks on the \textit{Data Manager} page. \textit{SafeSpeech} includes widely-used toxic content detection benchmarks: the SemEval2023 EDOS (\textit{sexism}) \cite{kirk-etal-2023-semeval}, HatEval (\textit{hateful}) \cite{basile-etal-2019-semeval}, AbusEval (\textit{abusive}) \cite{caselli-etal-2020-feel}, and OffensEval (\textit{offensive}) \cite{zampieri-etal-2019-semeval}.
The \textit{Data Manager} automatically parses the structure of the user-uploaded file based on column names, preprocessing them for message-level or conversation-level analyses 

\begin{figure*}[ht]
    \centering
    \includegraphics[width=\linewidth]{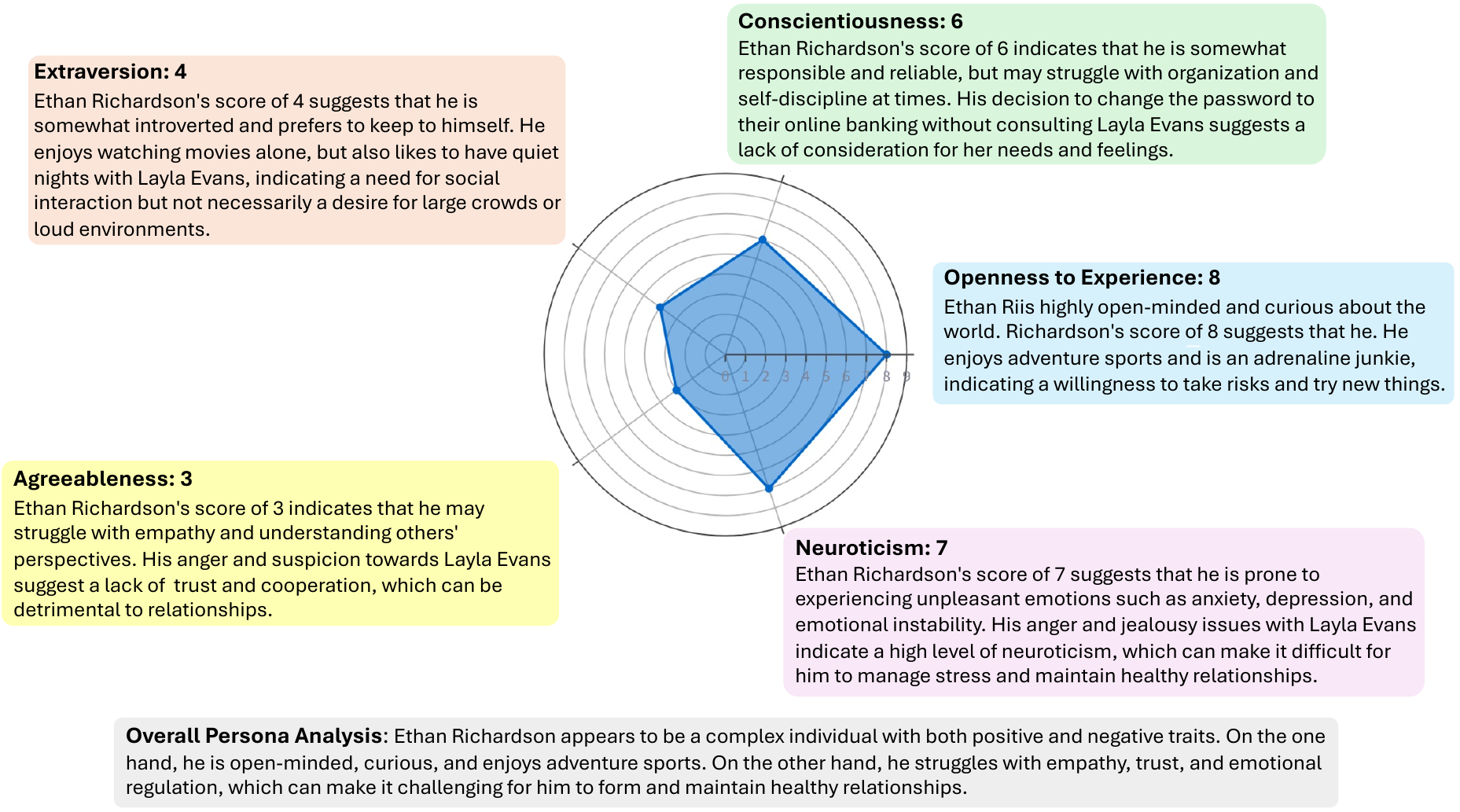}
    \caption{An example of persona analysis generated by Llama3.1 based on the conversation summary and the Big Five personality traits framework.}
    \label{fig:persona_example}
\end{figure*}

\subsubsection{Message Level Analyses} To investigate message-level data, \textit{SafeSpeech} provides state-of-the-art models and a customizable workflow in terms of labels and prompt templates. 
Users can select from multiple fine-tuned classification models to perform analyses across different levels of label granularity. E.g., in the default case of sexism detection, the platform allows classification on three levels of granularity: binary (sexist, non-sexist), four-class, and eleven-class subcategories \cite{kirk-etal-2023-semeval}.
Alongside fine-tuned models, users can enable the use of LLMs, which will automatically apply customized or predefined prompts to verify the classification results (See Appendix \ref{appendix:screenshoots} for screenshots).
Additionally, users can process to the \textit{Result Viewer} page for more visualization options.

\subsubsection{Conversations Level Analyses} To investigate conversation-level data, the platform provides several components to (i) highlight relevant excerpts of text, conduct (ii) topic-aware summarization, and (iii) persona profiling.

% PERPLEXITY
\noindent \textbf{Perplexity Gain} To identify the most relevant phrases determining the message classification, the platform relies on the analyses of the \textit{Perplexity Gain} \cite{suhara-alikaniotis-2024-source}. This statistical method quantify the contribution of individual phrases to the model's output (i.e., classification) by measuring the changes in perplexity when a particular sentence is removed from the input. The results are visualised as a heatmap over text, where darker shades denoted higher relevance. Figure \ref{fig:ppl_example} demonstrates an example on a synthetically generated conversation, where the LLM give a positive.

% SUMMARISATION
\noindent \textbf{Toxic-aware Speaker Summaries} The \textit{Summarization} component enables users to track the progression of multi-turn, multi-topic toxic conversations.
First, long conversations are visualised to the users into semantically coherent chunks, ensuring that recurrent topics discussed at different time intervals are logically grouped. For example, in a conversation where a topic like “workplace harassment” is initially discussed, interrupted by unrelated small talk, and then revisited later, the system identifies both segments as semantically related and groups them together. 
The summaries are generated for each segment, preserving each speaker's perspectives, and distinguishing contributions from different participants.
Additionally, the summarization process incorporated the message-level classifications as conditional inputs (\textit{toxic-aware summaries}). This is to highlight abusive messages within the condensed summaries and ensure that toxic or abusive content remains identifiable, a feature particularly relevant when dealing with subtle behaviours such as coercive control, backhanded compliments, or escalation patterns. 
To the user, intermediate results are visualised along with a progress bar, indicating the summarization is still in progress. Once processed, summaries generated for each semantically related chunk will be presented and grouped by participants to make it easier for users to navigate long conversations, understand the dynamics of the dialogue, and gain insights into how topics evolve (See Appendix \ref{appendix:screenshoots} for screenshots).

%PERSONA
\noindent \textbf{Persona Analysis} The \textit{Persona Analysis} component predicts the speaker's persona using the Big Five Personality Framework: Openness to Experience, Conscientiousness, Extraversion, Agreeableness, and Neuroticism \cite{costa2008revised}.  
Unlike general-purpose personality prediction approaches \cite{wen2024affective, han2023speaker, jcl132_personality_prediction}, this \textit{SafeSpeech}'s component incorporates toxic-aware conditional summaries to focus on relevant toxic contexts.
The module outputs numerical scores for each trait and explanatory justifications, linking linguistic features observed in the summaries to the predicted traits\footnote{The Persona Analysis module is intended for research purposes only. The generated profiles are exploratory and not definitive. Results rely on automated models that may reflect biases and inaccuracies inherent in the data or methods. They should not be used as a substitute for professional evaluation or in operational settings, but as a starting point to explore the limitations and potential of automated personality profiling in a controlled research environment.}. This information is presented via interactive radar plots, together with a detailed explanation of each trait, presenting a compressive analysis of each speaker's persona (Figure \ref{fig:persona_example}). For example, high scores in \textit{Neuroticism} might be associated with frequent emotional instability or escalation in toxic exchanges, while lower \textit{Agreeableness} could signal antagonistic tendencies in abusive or coercive contexts.

\subsubsection{AI Assistant} This module provides a chat-style interface where users can flexibly interact with the classified data via LLMs. 
This allows users to input prompts, request clarifications, or generate responses based on the classifier's prediction. 
The responses from the LLM are displayed via streaming, facilitating an iterative exploration process that helps refine the understanding of detected content. 
Users can also opt to apply predefined prompts or create custom ones, making this module adaptable for users with different levels of familiarity with prompt engineering.
The interaction history can then be conveniently downloaded.

%%%%%%%%%%%%%%%%%%%%%%%%%%%%%%%%%%%%%%%%%%%%%%%%%%
\subsection{Backend Implementation}
The backend of \textit{SafeSpeech} is designed to efficiently combine multiple data sources with fine-tuned neural classifiers and state-of-the-art LLMs.
It is based on \textit{FastAPI}\footnote{\url{https://github.com/fastapi/fastapi}}, a modern web framework that facilitates real-time communication between the frontend and backend, ensuring smooth data processing, model inference, and user interaction.  
The neural classifiers are implemented using \textit{PyTorch} and Huggingface's \textit{transformers}.
The LLMs are managed by \textit{Ollama} or via their publicly available APIs (e.g., OpenAI and Gemini).
Below, we describe how each module's methodologies, models, and tools are integrated into the backend.

\paragraph{AI Assistant} \textit{SafeSpeech} enables interactive exploration of toxic content detection by integrating LLMs through API-based communication with a conversation interface. Backend requests are managed via FastAPI, which routes user-defined or predefined prompts to the selected LLM for response generation. Open-weight LLMs, such as \textit{Llama3.1-70B-instruct}, are deployed using \textit{Ollama}, leveraging \textit{llama.cpp} for efficient model loading and inference. Users can flexibly choose LLMs based on model size, quantization levels, or available external APIs, such as OpenAI and Gemini.

\paragraph{Message Level Analysis} To produce message-level predictions, we utilize Transformer-based pre-trained language models, specifically \textit{DeBERTa-v3-large} \cite{2021deberta}, \textit{RoBERTa-Large} \cite{2019roberta}, and \textit{Mistral 7B} \cite{jiang23mistral}. 
These models are supervised fine-tuned on the respective task-specific datasets.
Moreover, \textit{SafeSpeech} integrates a voting-based ensemble model that combines the strengths of \textit{DeBERTa-v3-large}, \textit{RoBERTa-Large}, and \textit{Mistral 7B}, called \textit{M7-FE}, \cite{MFE2024} which outperforms existing models on sexist detection of English tweets.

\paragraph{Conversations Level Analysis} For the analyses of toxic conversations, we designed a backbone of LLMs, connecting the LLMs' APIs to the uploaded conversation files and to their message classifications at the conversation level via prompts.
This analysis is a prerequisite for \textit{Perplexity Gain}, \textit{Summarization}, and \textit{Persona Analysis}.

\paragraph{Perplexity Gain} The contribution of individual phrases to the model's generated results are computed via perplexity gain \cite{suhara-alikaniotis-2024-source}. 
It was originally proposed as a source identification approach for abstractive summarization, and it is based on the assumption that a language model should be more perplexed (i.e., less confident) to generate the same output (e.g., summary, or classification label) if an important sentence is removed.
In the analyses of conversations, when the LLM predicts that the input contains toxic contents, we adapted the perplexity gain to highlight the phrases that contribute the most to this conclusion.
In particular, the method first generates an output $Y$ from the complete input $X$, which consists of $N$ sentences ($s_1,...,s_i,...,s_N$).
Then, for each sentence $s_i$ in the input $X$, compute the perplexity of the original output $Y$ given an input $X$ in which the sentence $s_i$ is removed.
The difference between the initial perplexity and the perplexity after removing the sentence is used as a measure to evaluate how relevant the sentence $s_i$ is to the output $Y$.
The score is defined as:
\begin{equation}
    R(s_i,Y|X;\theta)=PPL(Y|X_{\setminus s_i};\theta)-PPL(Y|X;\theta)
\end{equation}
A heatmap visualizes perplexity gains, with darker shades indicating sentences most critical to the output. However, it is worth noticing that for negative predictions, perplexity gains are less informative as they rely on the entire conversation.

\paragraph{Toxic-aware Conditional Summarization} 
Our backend facilitates the generation of summaries for conversations among specific groups of participants.
It first identifies and organizes messages belonging to the same conversation by matching the list of dialogue participants.
Then, given that conversational flows often span multiple topics and temporal ranges, summarization is performed on utterances discussing the same topic \cite{Perg19}. This process requires segmenting lengthy conversations into semantically coherent chunks. To achieve this, we employ \textit{semantic chunking}\footnote{\url{https://docs.llamaindex.ai/en/stable/examples/node_parsers/semantic_chunking/}}, a method that adaptively determines segmentation points based on cosine similarity between sentence embeddings, generated using the \textit{sentence-transformers}\footnote{\url{https://huggingface.co/sentence-transformers}} library. We customized \textit{semantic chunking} to enhance its performance in processing conversational data.
Once the conversation is segmented, we employ an instructional summarization model, InstructDS \cite{wang-etal-2023-instructive}, to generate summaries for each chunk. InstructDS is a state-of-the-art instruction-following dialogue summarization model fine-tuned on Flan-T5-XL \cite{chung2022scalinginstructionfinetunedlanguagemodels}. 
The segmented content is input into the model with specific instructions to generate summaries considering messages previously categorized as toxic. 
The final summaries, categorized by segmented chunks, are transmitted to the frontend for display.

\begin{table*}[htb]
  \begin{center}
  \resizebox{0.9\textwidth}{!}{%
  \begin{tabular}{llr}
  \toprule
  \bf Dataset / Subtask & \bf Model & \bf Macro F1 \\ 
  \midrule 
  \multirow{9}{*}{EDOS Subtask A (Binary Sexism Detection)} 
  & \textit{HateBERT\_offeneval} & $0.53$ \\
  & \textit{HateBERT\_abuseval} & $0.59$ \\
  & \textit{HateBERT\_hateval} & $0.66$ \\
  & llama3.1-4bit (in-context learning) & $0.60$ \\
  & \textit{NLP-LTU/bertweet}  & $0.86$ \\
  & PingAnLifeInsurance \cite{zhou-2023-pinganlifeinsurance} & $0.87$ \\
  & \textit{\textbf{M7-FE (Our)}} & $\textbf{0.88}$ \\
  \midrule
  
  \multirow{3}{*}{EDOS Subtask B (Category Detection, 4 Classes)} 
  & \textit{ISEGURA/roberta} & $0.58$ \\
  & PingAnLifeInsurance \cite{zhou-2023-pinganlifeinsurance} & $0.72$ \\
  & \textit{\textbf{M7-FE (Our)}} & $\textbf{0.72}$ \\
  \midrule
  
  \multirow{3}{*}{EDOS Subtask C (Vector Detection, 11 Classes)} 
  & \textit{ISEGURA/roberta} & $0.43$ \\
  & PingAnLifeInsurance \cite{zhou-2023-pinganlifeinsurance} & $0.56$ \\
  & \textit{\textbf{M7-FE (Our)}} & $\textbf{0.60}$ \\
  \hline\hline
  
  \multirow{5}{*}{OffensEval 2019}
  & \textit{M7-FE} (Our) & $0.48$ \\
  \arrayrulecolor[gray]{0.75}\cline{2-3}
  \arrayrulecolor{black}
  & Llama3.1-4bit (in-context learning) & $0.67$ \\
  & BERT  & $0.80$ \\
  & HateBERT & $\textbf{0.81}$ \\
  \hline\hline
  
  \multirow{5}{*}{AbusEval}
  & \textit{M7-FE} (Our) & $0.49$ \\
  \arrayrulecolor[gray]{0.75}\cline{2-3}
  \arrayrulecolor{black}
  & Llama3.1-4bit (in-context learning) & $0.65$ \\
  & BERT  & $0.73$ \\
  & HateBERT & $\textbf{0.77}$ \\
    \hline\hline
  
  \multirow{5}{*}{HatEval}  
  & \textit{M7-FE} (Our) & $0.51$ \\
  \arrayrulecolor[gray]{0.75}\cline{2-3}
  \arrayrulecolor{black}
  & Llama3.1-4bit (in-context learning) & $\textbf{0.59}$ \\
  & BERT  & $0.48$ \\
  & HateBERT  & $0.52$ \\
  \bottomrule
  \end{tabular}}
  \end{center}
  \caption{The test performance of the available models on the benchmark datasets on \textit{SafeSpeech}. BERT and HateBERT results are obtained from \citet{caselli-etal-2021-hatebert}. \vspace{-12pt}}
  \label{merged_table}
\end{table*}

\paragraph{Summary-based Persona Analysis} 
This module leverages dialogue summaries to predict speaker personas based on the Big Five personality traits framework \cite{costa2008revised}. The backend of this module involves three key components: summary concatenation, persona prediction, and output generation.
First, all relevant toxic-aware summaries generated from prior stages are concatenated. This provides a comprehensive context for a speaker's linguistic style, behavioural patterns, and interpersonal dynamic. 
By leveraging summaries instead of lengthy raw messages, the method not only minimizes information loss on critical toxic-related patterns or utterances, but also ensures computational efficiency. 
The concatenated summaries are subsequently processed by a persona prediction model, instructed via prompt to identify linguistic and behavioural patterns based on the Big Five personality traits theory.
Detailed prompts are shown in the Appendix \ref{persona_prompt_table}. 
Due to data availability constraints, we assume for simplicity and to reduce sparsity, a consistent persona across all messages for each speaker. This assumption enables the generation of a unified and cohesive persona profile for each individual within the dialogue.

\section{Evaluation}
\subsection{Quantitative Analysis}
We evaluate the performance of \textit{SafeSpeech}’s message-level classification models across widely-used benchmarks for toxic content detection.
Table \ref{merged_table} reports results on benchmark datasets including EDOS - SemEval 2023 \cite{kirk-etal-2023-semeval}, OffensEval 2019, AbusEval, and HatEval.

For sexism detection (EDOS), we evaluate classifiers on three levels of granularity: binary detection (Subtask A), category-level classification (Subtask B), and fine-grained vector-level classification (Subtask C). \textit{SafeSpeech} integrates state-of-the-art models, including our ensemble method \textbf{M7-FE} (Mistral-7B Fallback Ensemble). The results demonstrate that \textbf{M7-FE} achieves top performance across all subtasks, improving over the winning system \cite{zhou-2023-pinganlifeinsurance} by 1\% F1 on Subtask A and 4\% F1 on Subtask C. On Subtask B, \textbf{M7-FE} matches the best reported performance while outperforming publicly available baselines such as \textit{ISEGURA/roberta}\footnote{\url{ISEGURA/roberta-base_edos_b}} by $14\%$ F1. \textbf{M7-FE} outperforms \textit{NLP-LTU/bertweet}\footnote{\href{https://huggingface.co/NLP-LTU/bertweet-large-sexism-detector}{NLP-LTU/bertweet-large-sexism-detector}} by $2\%$ and prompting with the 4-bit quantised version of Llama3.1.

For hate speech, abusive speech, and offensive speech classification, we evaluate our models on the OffensEval, AbusEval, and HatEval benchmarks. Notably, \textit{HateBERT} \cite{caselli-etal-2021-hatebert}, a model specifically designed for this domain, achieves top-ranking performance on the OffensEval 2019, AbusEval, and HatEval datasets. Specifically, \textit{HateBERT} demonstrates the highest F1 scores on OffensEval 2019 and AbusEval when compared to baselines such as vanilla BERT and a 4-bit quantized Llama 3.1 model. Conversely, the 4-bit Llama 3.1 model exhibits superior performance on the HatEval dataset. Additionally, we assess the M7-FE model, trained on the EDOS dataset, to evaluate its performance in slightly out-of-domain settings across the same benchmarks. While M7-FE does not surpass the performance of Llama 3.1 or supervised fine-tuned BERT models, it achieves comparable results on HatEval, suggesting a significant overlap between sexism and hate speech as represented in these datasets.

\section{Conclusion and Future Work}

\textit{SafeSpeech} is the first platform to provide tools for conveniently analysing hate speech or violence against women and girls in a conversational format.
It connects with state-of-the-art supervised fine-tuned models and large language models, offering a comprehensive and interactive interface for researchers to test their new ideas without the need for coding.
In future work, we will try to expand the platform and make it easy to use for the general public, making it an accessible tool for social media moderation.

\section*{Limitations}
\paragraph{Bias Analysis} Our platform has not analyzed the bias analysis of LLMs on detecting toxic content.
Social media content makes up for a big portion of the pre-training data of the LLMs.
Although most of the LLMs include filtering about toxic content when selecting their training data, there may still be implicit bias in the training data.

\paragraph{Model Transparency} Although we try to generate explanations and perplexity gains for the classifications, the transparency of the classifiers and LLMs remains an area for further improvement.
The generated explanations are external references instead of a direct interpretation of the classifiers' internal mechanism.
The perplexity gain was originally designed for abstraction summarization, and its effect on toxic content detection still requires more examination.
The generated personality traits are predicted merely based on the context correlation between the inputs and the prompt, it has not been clinically validated and the output should not be used for professional diagnosis.
Despite all these limitations, we still think the \textit{SafeSpeech} platform would promote the exploration of toxic content detection by providing convenient access to these functions.

\paragraph{Model Variety} The demo currently provides a limited selection of models due to computational constraints.
However, users can load additional models from the HuggingFace Model Hub, subject to size restrictions.
Additionally, users can also integrate the use of external LLMs via their API keys.

\section*{Ethics Statement}

\paragraph{Data Security and Privacy}
The current platform is a prototypical demo which has not implemented extensive cybersecurity measures.
The benchmarking data used in our platform are from publicly available datasets which do not contain confidential data.
We recommend that users only upload non-sensitive data that can be publicly shared for research purposes.

\paragraph{The Risks in Monitoring Hate Speech}
This paper focuses on toxic content detection and analysis, including sexism, racism, and xenophobic language etc.
However, the definitions of such content are hugely controversial and are changing rapidly over time.
The actions of monitoring and censoring hate speech without accurate and socially agreed definitions might bring risks of limiting the freedom of speech.
Moreover, the training of LLMs and other neural models might already incorporate social biases which might affect the classification and explanation of the detection system.
An additional risk is that the monitoring and censoring of toxic content might shift the form of such content and make understanding it more challenging.

\section*{Acknowledgements}
Several authors were partially supported by the Police Science, Technology, Analysis, and Research (STAR) Fund 2022–23, funded by the National Police Chiefs' Council (NPCC), in collaboration with the Forensic Capability Network (FCN). 
Xingwei Tan was supported by the Warwick Chancellor's International Scholarship.
This work was conducted on the Sulis Tier-2 HPC platform hosted by the Scientific Computing Research Technology Platform at the University of Warwick. Sulis is funded by EPSRC Grant EP/T022108/1 and the HPC Midlands+ consortium.

\bibliography{anthology,custom}
\bibliographystyle{acl_natbib}

\newpage
\clearpage
\appendix

\section{Appendix}
\label{sec:appendix}

\subsection{Experiment Results on EXIST}
CLEF 2024 sEXism Identification in Social neTworks (EXIST)\footnote{\href{http://nlp.uned.es/exist2024/}{EXIST official website}} is another competition on sexism content detection.
Mistral-7B Fallback Ensemble (M7-FE) \cite{MFE2024}, which our team members develop, ranks first among $68$ submitted models on English test data with an F1 of $0.76$.
Dual-Transformer Fusion Networks (DTFN) \cite{MFE2024}, which is also developed by our team members, ranks second among the $68$.
I2C-UHU \cite{DBLP:conf/clef/Guerrero-Garcia24} ranks third in the leaderboard.
Table \ref{merged_table} shows the official results\footnote{\href{https://docs.google.com/spreadsheets/d/1P63RjIEnnrRFqQiXsHLAhS7nKA0dhrny/edit?gid=517615678\#gid=517615678}{EXIST task 1 English leaderboard}} of the test set evaluation.
The ICM metric, proposed by \cite{amigo-delgado-2022-evaluating}, is based on information theory and measures the similarity between system classifications and gold standard labels.
The normalized version, ICM-Hard Norm, takes the label imbalances into account.
Higher values of ICM and ICM-Hard Norm indicate a stronger alignment between system outputs and the ground truth.

\begin{table}[htb]
  \begin{center}
  \resizebox{\columnwidth}{!}{%{
  \begin{tabular}{lccc}
  \toprule
  \bf Model & ICM-HARD & ICM-Hard Norm & F1 \\ 
  \midrule 
  I2C-UHU  & $0.58$ & $0.80$ & $0.76$ \\
  DTFN \textit{(Our)}  & $0.60$ & $0.80$ & $0.75$\\
  \textbf{M7-FE} \textit{(Our)} & $\textbf{0.62}$ & $\textbf{0.82}$ & $\textbf{0.76}$\\
  \bottomrule
  \end{tabular}}
  \end{center}
  \caption{The performance of the models on the EXIST task 1 English dataset. }
  \label{exist_table}
\end{table}

\subsection{Prompt Templates}
\label{appendix:prompts}

\begin{table}[H]
  \centering
  \begin{tabularx}{\linewidth}{|X|}
    \hline
    \rowcolor{rowcolor} Default VAWG content detection prompt \\ 
    \hline
    Instruction: Does the given conversation contain content about violence against women and girls? \\

    VAWG definitions: \\

    The followings are types of violence against women and girls: \\
    1) Domestic Abuse; 2) stalking and harassment; 3) controlling or coercive behaviour; 4) child abuse; 5) rape; 6) honour based abuse; 7) forced marriage; 8) human trafficking, smuggling and slavery; 9) female genital mutilation; 10) prostitution. \\

    Conversation: \\
    """\{conversation\}""" \\
    \hline
  \end{tabularx}
  \caption{An example of the classification prompt when the toxic type is violence against women and girls.}
  \label{classification_prompt_table_vawg}
\end{table}
%%%%%%%%%%%%%%%%%%%%%%%%%%%%%%%%%%%%%%%%%%%

%%%%%%%%%%%%%%%%%%%%%%%%%%%%%%%%%%%%%%%%%%%
\begin{table}[H]
  \centering
  \begin{tabularx}{\linewidth}{|X|}
    \hline
    \rowcolor{rowcolor} Default sexism content detection prompt \\ 
    \hline
    Instruction: Does the given conversation contain content about sexism? \\

    Sexism definitions: \\

    The followings are types of sexism: \\
    1) threats; 2) derogation; 3) animosity; 4) prejudiced discussion. The types can be further divided into 11 categories: Threats of harm; Incitement and encouragement of harm; Descriptive attacks; Aggressive and emotive attacks; Dehumanising attacks and overt sexual objectification; Causal use of gendered slurs, profanities and insults; Immutable gender differences and gender stereotypes; Backhanded gendered compliments; Condescending explanations or unwelcome advice; Supporting mistreatment of individual women; Supporting systemic discrimination against women as a group. \\

    Conversation: \\
    """\{conversation\}""" \\
    \hline
  \end{tabularx}
  \caption{An example of the classification prompt when the toxic type is sexism.}
  \label{classification_prompt_table_sexism}
\end{table}
%%%%%%%%%%%%%%%%%%%%%%%%%%%%%%%%%%%%%%%%%%%

%%%%%%%%%%%%%%%%%%%%%%%%%%%%%%%%%%%%%%%%%%%
\begin{table*}
  \begin{center}
  \resizebox{\textwidth}{!}{%
  \begin{tabularx}{\textwidth}{|X|}
\hline
\rowcolor{rowcolor} Default persona generation prompt \\ 
  \hline
  
You will be given a concatenated summary of all the chat messages that [\textit{speaker}] participated in. Your task is to read the given summary and provide numeric Big Five Personality prediction scores for [\textit{speaker}].

\vspace{\baselineskip}

The Big Five personality traits are a widely recognized model for describing human personality. The traits are:

    Openness to Experience: Reflects the degree of intellectual curiosity, creativity, and a preference for novelty and variety.
    Conscientiousness: Indicates a tendency for self-discipline, carefulness, and a goal-oriented behaviour.
    
    Extraversion: Involves energy, positive emotions, and a tendency to seek stimulation and the company of others.
    
    Agreeableness: Represents a tendency to be compassionate and cooperative towards others.
    
    Neuroticism: Indicates a tendency to experience unpleasant emotions easily, such as anxiety, depression, and emotional instability.
    
\vspace{\baselineskip}

Give integer prediction scores for each trait in Big-Five Persona for [\textit{speaker}], based on the chat message summary that will be given. Generate your response following these steps:

1. Read the summary thoroughly, especially pay attention to those content related to toxic behaviour.

2. Generate list of INTEGERS (from 1 to 10) as Big Five Personality prediction scores, each associated with a trait in order, separated by commas, wrapping with square brackets. 

3. Double check to ensure that only 5 INTEGER NUMBERS are provided at the BEGINNING of your response. 

4. For each trait, provide a more detailed explanation for the score predicted.

5. Generate 1-2 sentences to give an overall persona analysis for [\textit{speaker}].

\vspace{\baselineskip}

Chat Messages Summary:

[\textit{summary}]

\vspace{\baselineskip}

Generate the respond in the following format: 

"[score 1, score 2, score 3, score 4, score 5]

**Openness to Experience**: <explanation>

**Conscientiousness**: <explanation>

**Extraversion**: <explanation>

**Agreeableness**: <explanation>

**Neuroticism**: <explanation>

**Overall Persona Analysis**: <analysis>"

\\
  \hline
  \end{tabularx}}
  \end{center}
  \caption{An example of the persona generation prompt for big five personality trait generation.}
  \label{persona_prompt_table}
\end{table*}
%%%%%%%%%%%%%%%%%%%%%%%%%%%%%%%%%%%%%%%%%%%

%%%%%%%%%%%%%%%%%%%%%%%%%%%%%%%%%%%%%%%%%%%
\begin{table}[H]
  \centering
  \begin{tabularx}{\linewidth}{|X|}
    \hline
    \rowcolor{rowcolor} Default sexist content explanation prompt \\ 
    \hline
    Based on the analysis of a classifier, the conversation below contains sexist messages. The details are as follows: \\

    \{labels\} \\

    Based on the prediction, could you explain why? Is there any other sexist content in the conversation? \\

    Conversation: \\
    """\{conversation\}""" \\
    \hline
  \end{tabularx}
  \caption{An example of the explanation prompt when the toxic type is violence against women and girls.}
  \label{explanation_prompt_table}
\end{table}
%%%%%%%%%%%%%%%%%%%%%%%%%%%%%%%%%%%%%%%%%%%

\subsection{Example Conversation}

\begin{figure*}[t]
  \centering
  \includegraphics[width=.9\textwidth]{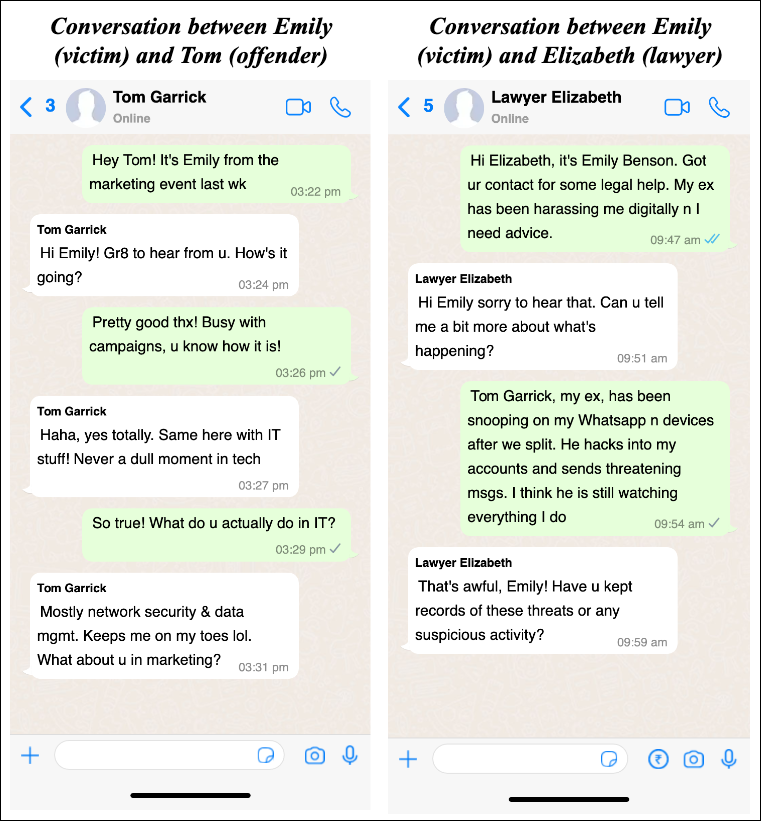}
  \caption{An example conversation about harassment.}
  \label{fig:dv-conv1}
\end{figure*}

\begin{figure*}[t]
  \centering
  \includegraphics[width=.9\textwidth]{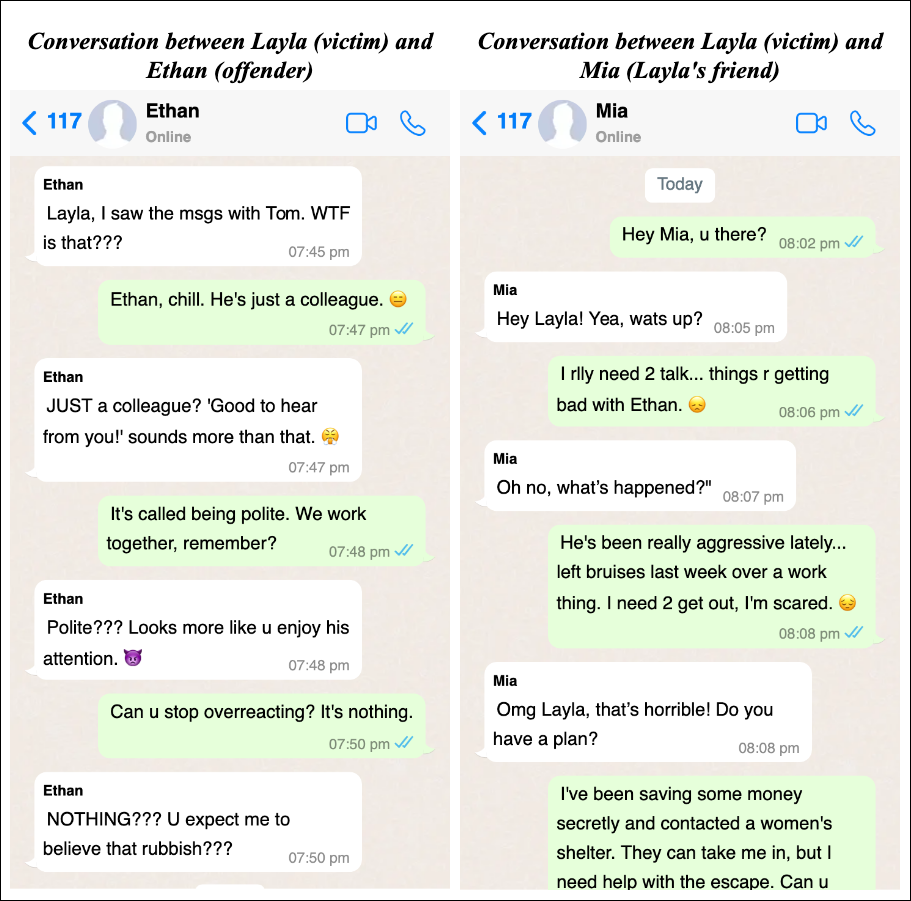}
  \caption{An example conversation involving controlling and coercive behaviours.}
  \label{fig:dv-conv2}
\end{figure*}

Figure \ref{fig:dv-conv1} and Figure \ref{fig:dv-conv2} show examples related to violence against women and girls. Such cases have become the main focuses of the Forensic Capability Network.

\subsection{\textit{SafeSpeech} Interface Screenshots}
\label{appendix:screenshoots}
In this section, we provide a detailed overview of the main modules of \textit{SafeSpeech} through interface screenshots. These images illustrate the platform's key functionalities.

\begin{figure*}[htb]
    \centering
    \includegraphics[width=0.7\textwidth]{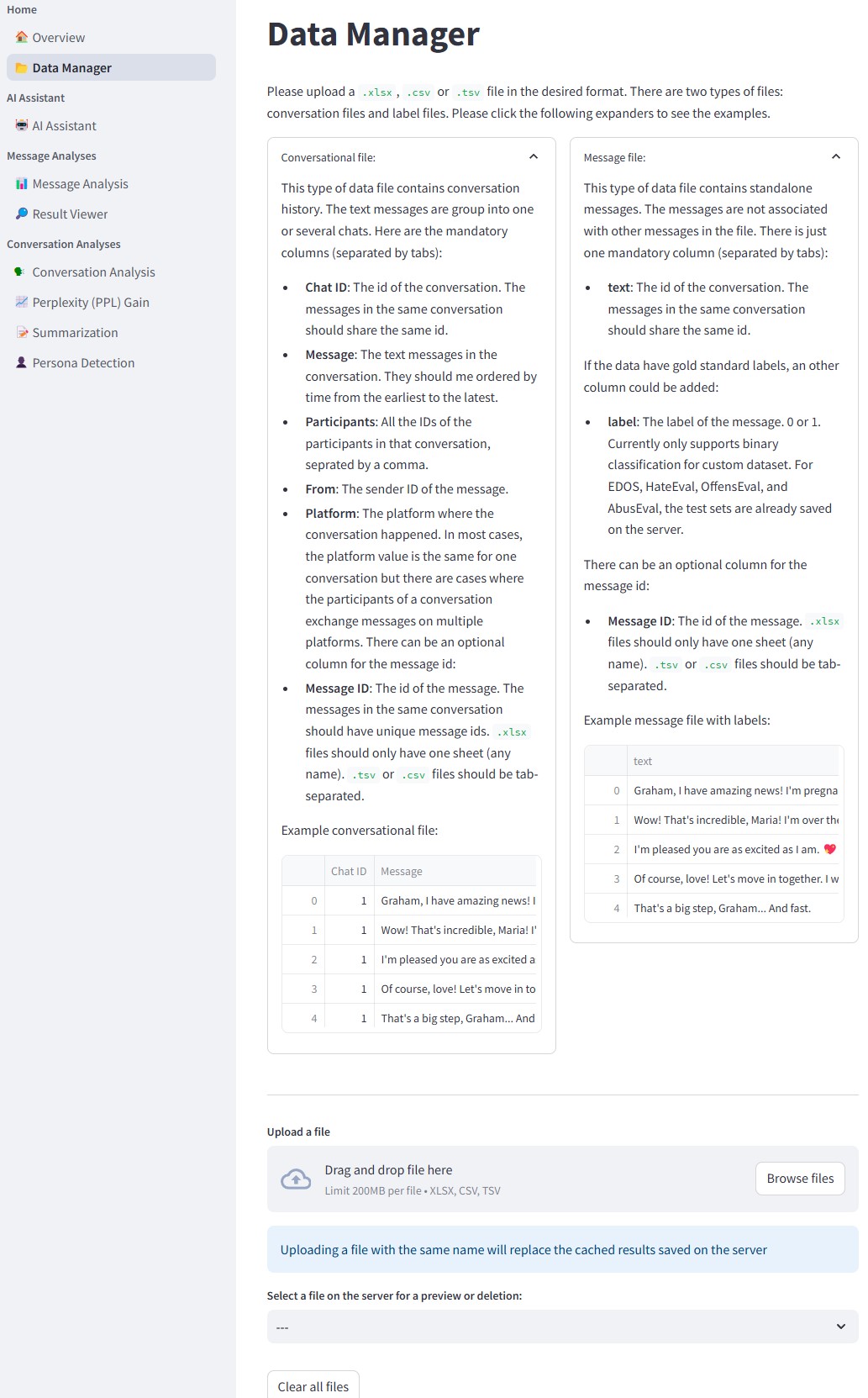}
    \caption{\textbf{A screenshot of the data manager page.} Users can upload their own data and preview the benchmark datasets on the platform. They can use the \textit{delete} button to delete their own files.}
    \label{fig:file_manager}
\end{figure*}

\begin{figure*}
    \centering
    \includegraphics[width=0.9\textwidth]{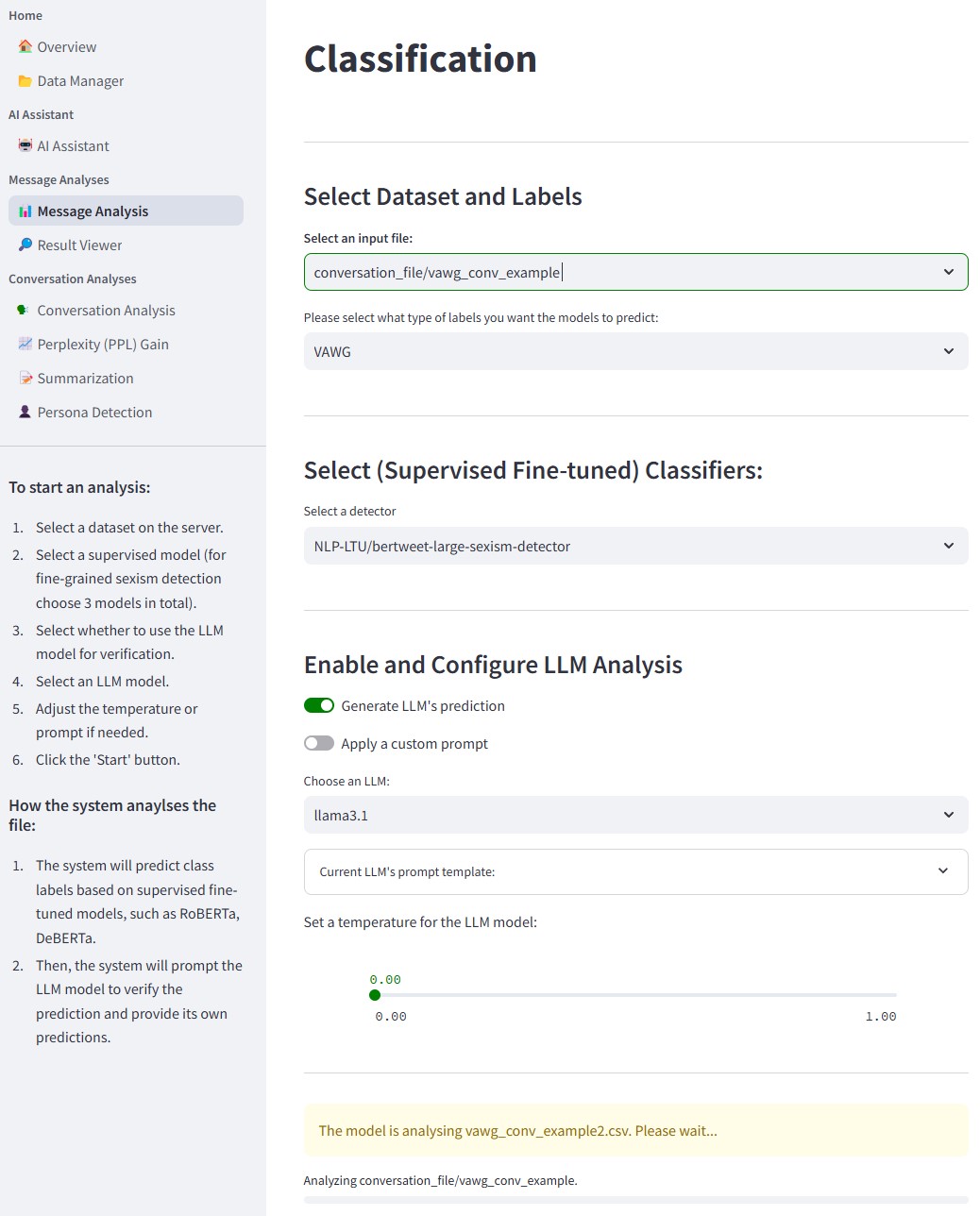}
    \caption{\textbf{A screenshot of the classification page.} Users can select the classification models and change the hyperparameters. After completing the setup, they can hit the \textit{start} button to start the analysis.}
    \label{fig:classification_1}
\end{figure*}

\begin{figure*}
    \centering
    \includegraphics[width=0.9\textwidth]{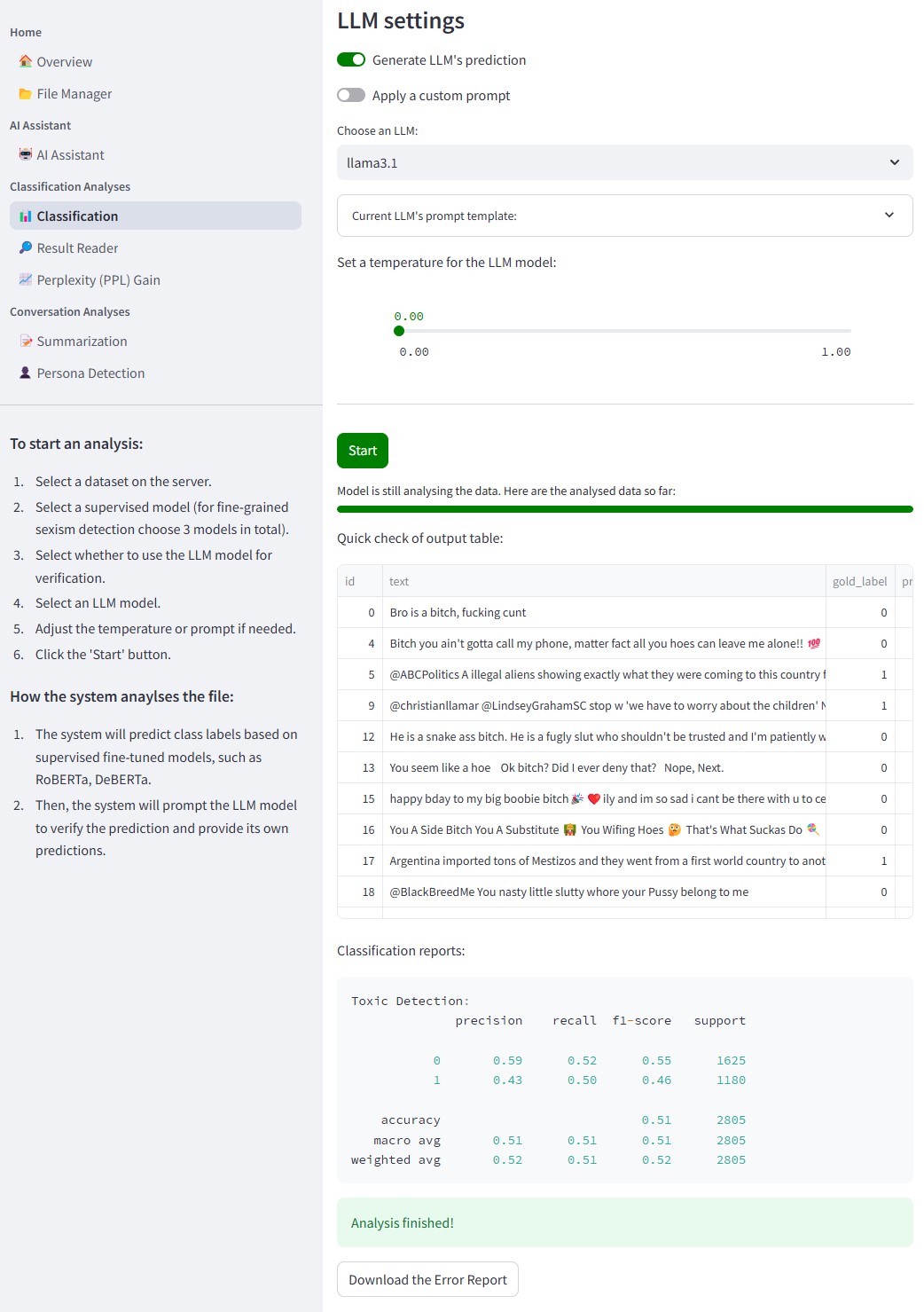}
    \caption{\textbf{A screenshot of the classification page after finishing the analysis.} The users can preview and download the output file. Our system will show a classification report if the selected file contains gold labels.}
    \label{fig:classification_2}
\end{figure*}

\begin{figure*}[htb]
    \centering
    \includegraphics[width=0.9\textwidth]{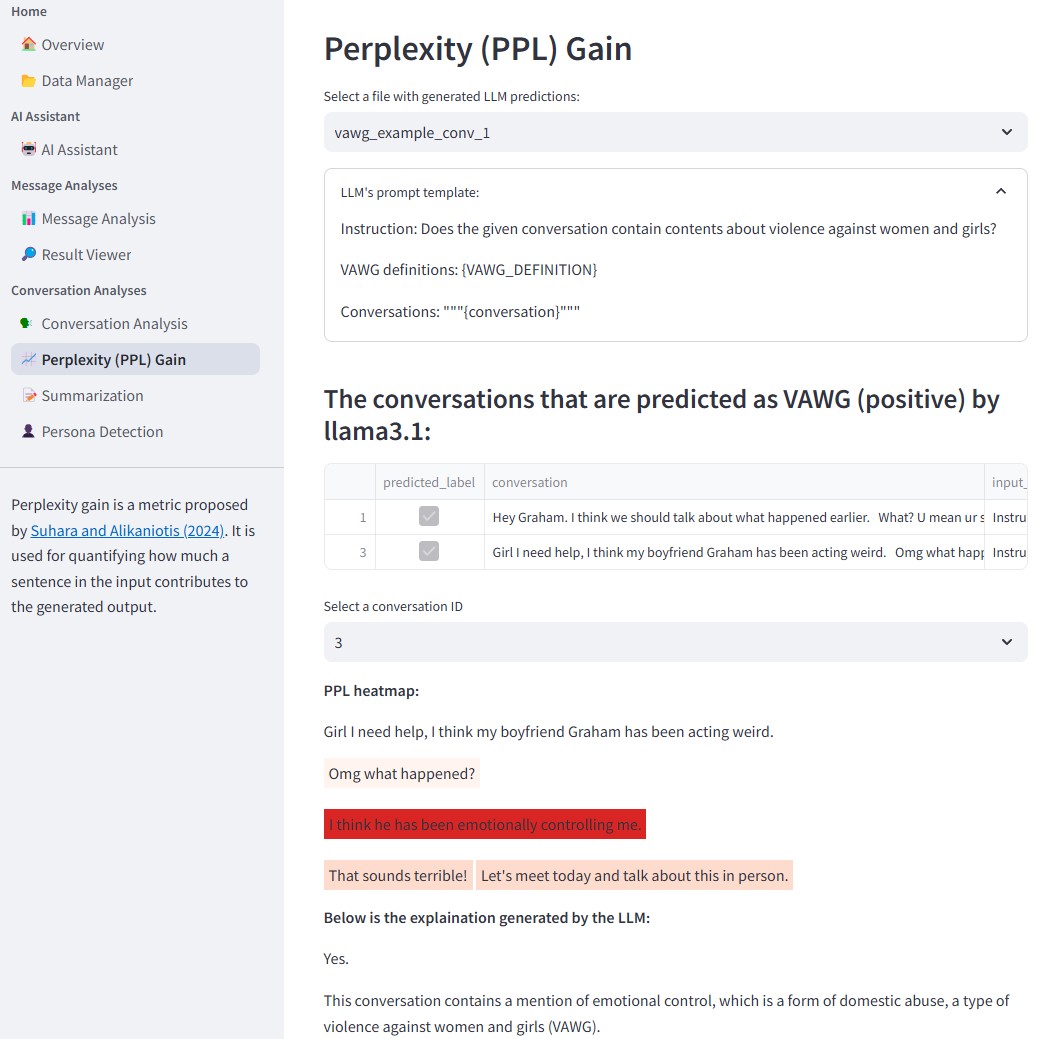}
    \caption{\textbf{A screenshot of the perplexity gain analysis page.} After choosing a conversation from the selected file, our system generates and then displays a heatmap of perplexity gains on the messages in that conversation.}
    \label{fig:ppl_gain}
\end{figure*}

\begin{figure*}
    \centering
    \includegraphics[width=0.9\textwidth]{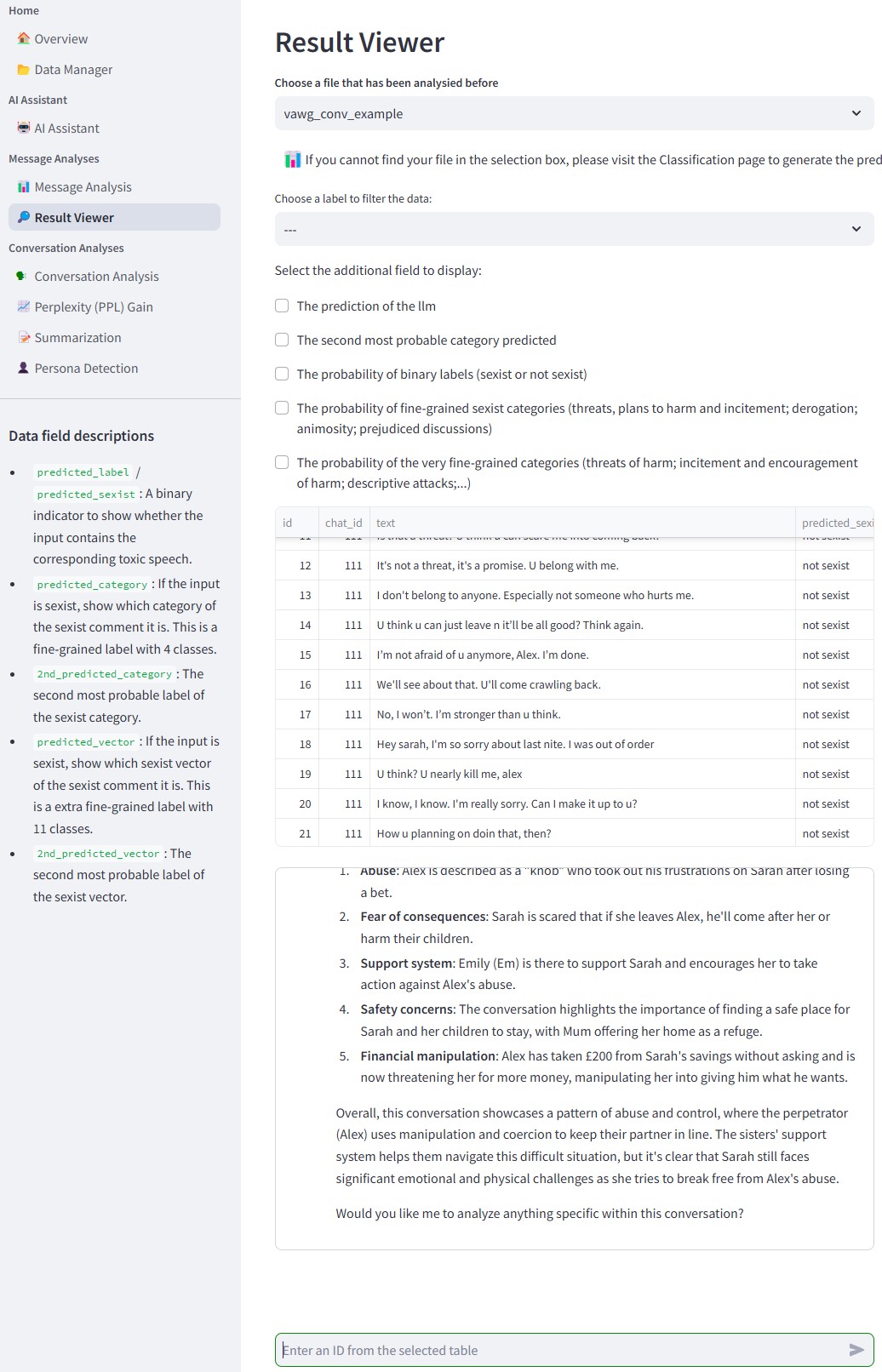}
    \caption{\textbf{A screenshot of the result viewer page.} We provide a reader on our platform to let users view the result files without downloading them. Users can choose the data they would like to view, such as the $2^{\textrm{nd}}$ most probable labels. We also integrate a conversation interface on this page which can provide explanations when the user indicate an index in the file.}
    \label{fig:result_reader}
\end{figure*}

\begin{figure*}[htb]
    \centering
    \includegraphics[width=0.9\textwidth]{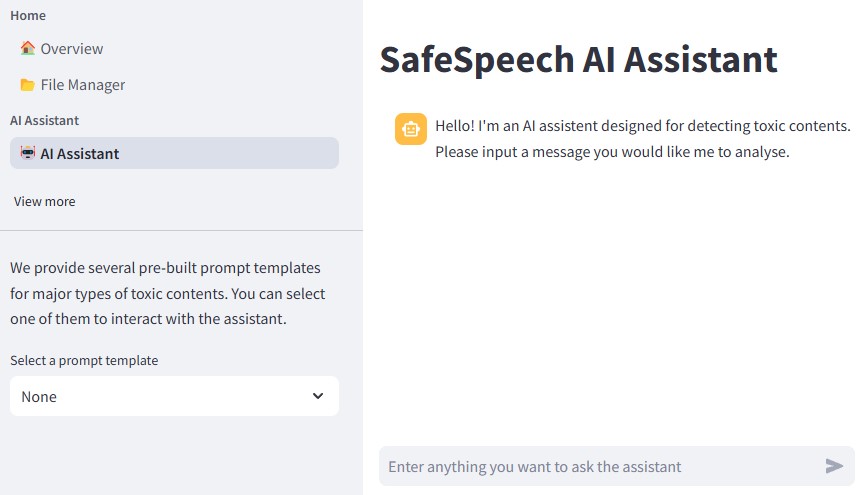}
    \caption{\textbf{A screenshot of the AI assistant page.} We provide a conversation interface, which is aided with carefully-designed prompts and classification labels, for interactive detection of toxic content. Users can also input custom prompts for detection. They can also ask follow-up questions.}
    \label{fig:ai_assistant}
\end{figure*}

\begin{figure*}[htb]
    \centering
    \includegraphics[width=0.9\textwidth]{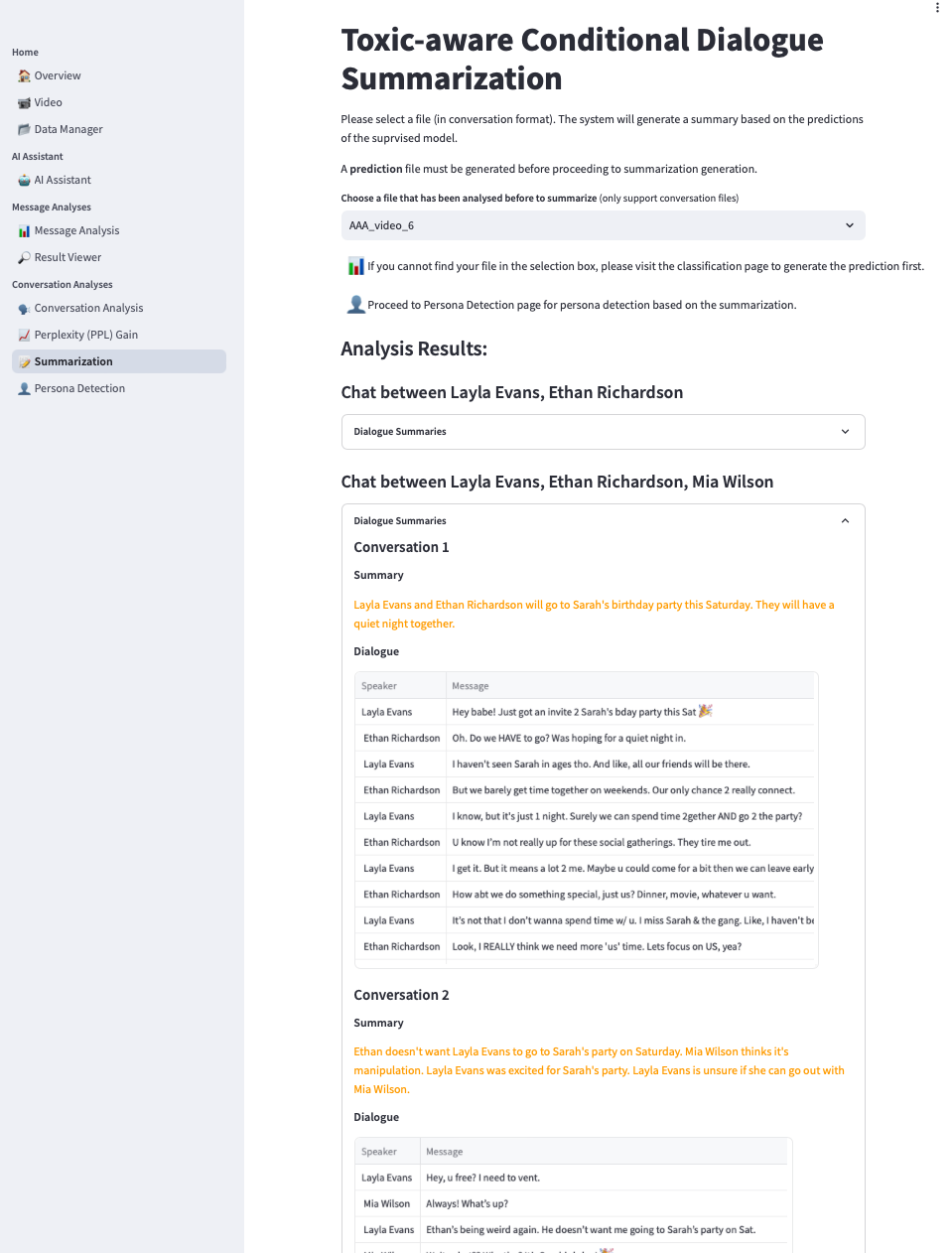}
    \caption{\textbf{A screenshot of the Summarization page.} The interface displays conversation summaries segmented into logical chunks based on topics. Users can view concise, toxic-aware summaries that highlight relevant toxic or abusive content. The progress bar ensures users can track the status of the summarization process.}
    \label{fig:sum}
\end{figure*}

\begin{figure*}[htb]
    \centering
    \includegraphics[width=0.9\textwidth]{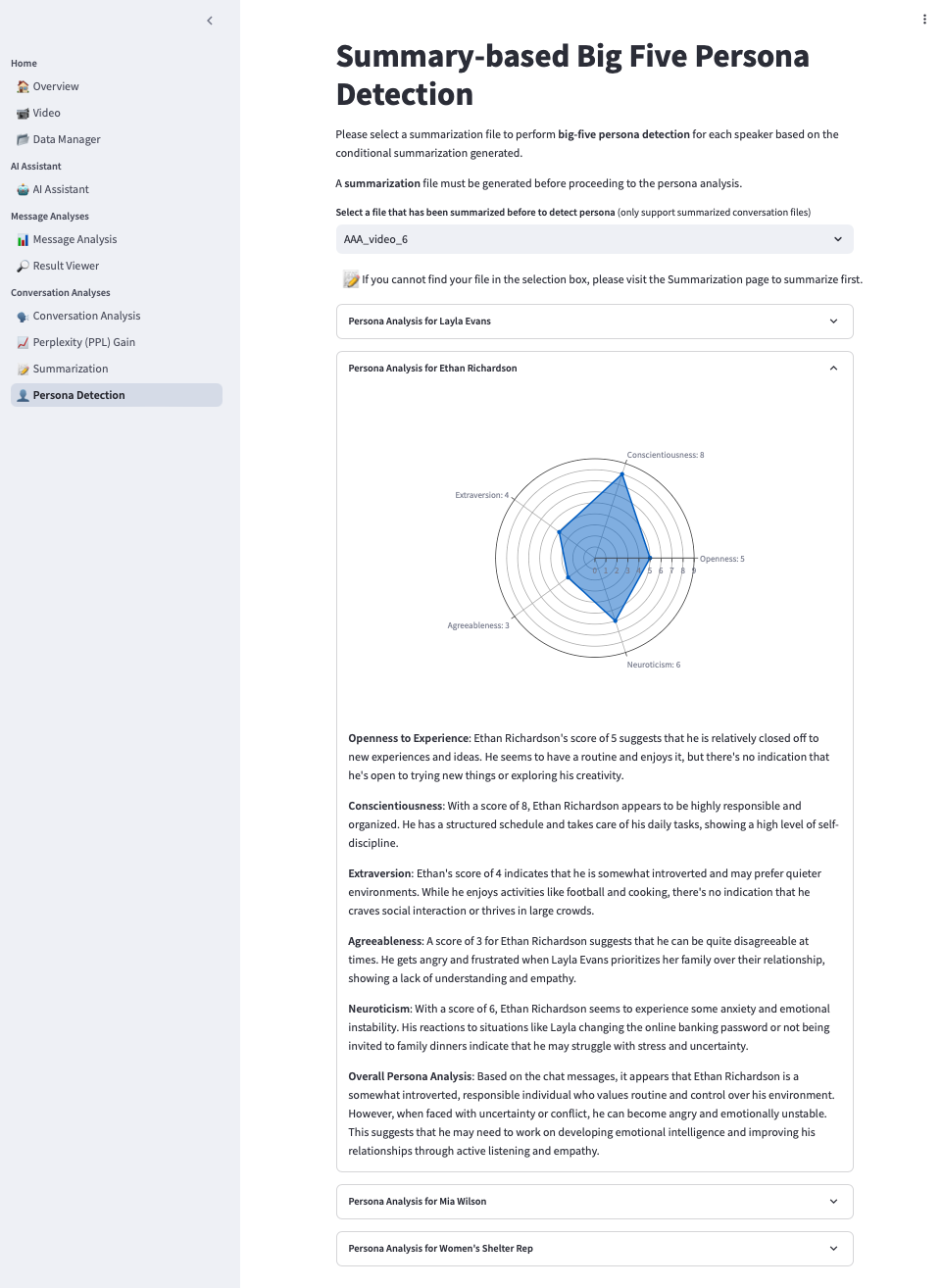}
    \caption{\textbf{A screenshot of the Persona Detection page.} This interface provides a visual representation of persona analysis results. Users can explore personality trait scores using an interactive radar plot and read detailed explanations for each trait. The clear layout helps users quickly interpret the predicted persona of conversation participants.}
    \label{fig:persona}
\end{figure*}

\end{document}